\documentclass{article}

    \usepackage[final]{tackling_climate_workshop_style}

\usepackage[utf8]{inputenc} % allow utf-8 input
\usepackage[T1]{fontenc}    % use 8-bit T1 fonts
\usepackage{hyperref}       % hyperlinks
\usepackage{url}            % simple URL typesetting
\usepackage{booktabs}       % professional-quality tables
\usepackage{amsfonts}       % blackboard math symbols
\usepackage{nicefrac}       % compact symbols for 1/2, etc.
\usepackage{microtype}      % microtypography
\usepackage{graphicx} % jnote LOADED THIS FOR FIGURES
\usepackage{float} % jnote added this for appendix tables bc they were randomly floating

\title{A Graph Neural Network Approach for Localized and High-Resolution Temperature Forecasting}

\author{
  Joud El-Shawa$^{1,2}$ \quad 
  Elham Bagheri$^{1,2}$ \quad 
  Sedef Akinli Kocak$^{1}$ \quad 
  Yalda Mohsenzadeh$^{1,2*}$ \\
  $^1$Vector Institute for Artificial Intelligence, Toronto, Canada   \\
  $^2$Western University, London, Canada \\
  $^{*}$\texttt{ymohsenz@uwo.ca} \\
}

\begin{document}

\maketitle

\begin{abstract} 
  % must be limited to one paragraph. 
  Heatwaves are intensifying worldwide and are among the deadliest weather disasters. The burden falls disproportionately on marginalized populations and the Global South, where under-resourced health systems, exposure to urban heat islands, and the lack of adaptive infrastructure amplify risks. Yet current numerical weather prediction models often fail to capture micro-scale extremes, leaving the most vulnerable excluded from timely early warnings. We present a Graph Neural Network framework for localized, high-resolution temperature forecasting. By leveraging spatial learning and efficient computation, our approach generates forecasts at multiple horizons, up to 48 hours. For Southwestern Ontario, Canada, the model captures temperature patterns with a mean MAE of 1.93\textdegree C across 1–48h forecasts and MAE@48h of 2.93\textdegree C, evaluated using 24h input windows on the largest region. While demonstrated here in a data-rich context, this work lays the foundation for transfer learning approaches that could enable localized, equitable forecasts in data-limited regions of the Global South.
\end{abstract}

\section{Introduction}
A global study estimated that from 2000-2019 approximately 489{,}000 heat-related deaths occurred each year \cite{zhao2021global}, a figure also highlighted by the World Health Organization \cite{WHO2019heatwaves}, and one that has likely risen as the climate crisis accelerates. Heatwaves threaten health, ecosystems, and economies worldwide. Their impacts are uneven: low-income, racialized, and Global South communities are most exposed, despite minimal contribution to emissions \cite{parsons2024climate, deivanayagam2023envisioning}. Factors such as poor housing insulation, limited cooling infrastructure, and underfunded public health amplify vulnerability \cite{smith2022climate,hansen2013vulnerability}. At the national scale, low- and middle-income countries face severe risks due to lower adaptive capacity and constrained resources \cite{ziegler2017climate}.

Unfortunately, current operational weather forecasts often lack the granularity and lead time needed to adequately warn and protect local communities. Existing forecast systems typically operate at 10-30~km scales, smoothing over urban heat islands or neighborhood ``hot spots.'' This leads to systematic underestimation of heat risk in precisely those marginalized areas most in need of targeted interventions \cite{putsoane2024extreme}. Addressing this inequity requires forecasting systems that are high-resolution, adaptive, and accessible across diverse contexts, from urban heat islands in megacities to rural, resource-limited regions.

%\section{Background and related work}
Traditional numerical weather prediction (NWP) models are computationally costly, limited in resolution, and rely on parameterizations that often miss land-atmosphere feedbacks critical for heat extremes \cite{bauer2015quiet, seneviratne2010investigating}. Recent machine learning advances demonstrate alternatives: GraphCast achieves skillful global forecasts at 28~km resolution \cite{lam2023graphcast}, FourCastNet uses Fourier operators for global predictions at a similar resolution \cite{pathak2022fourcastnet}, and neural models improve extreme heat anomaly forecasts \cite{lopez2023global}.
Most approaches, however, emphasize global scales. Localized prediction remains underdeveloped, despite its importance for frontline adaptation. Li et al. \cite{li2023regional} showed that Graph Neural Networks (GNNs) can classify regional heatwave events with high accuracy, underscoring their potential for fine-scale, actionable forecasting. Our work extends this logic: using GNNs to produce continuous, high-resolution temperature forecasts that can be post-processed to identify heatwaves. This provides a step toward developing practical, fine-scale forecasting frameworks that can eventually be adapted to data-limited or vulnerable contexts in the Global South.

\section{Methods}

\paragraph{Data.} Our study leverages the National Oceanic and Atmospheric Administration (NOAA) Unrestricted Mesoscale Analysis (URMA) dataset as the primary source of training and evaluation data \cite{noaa_rtma_urma}. URMA provides high-resolution (2.5~km, hourly), gridded analyses of surface meteorological variables, such as temperature, winds, pressure, and elevation, which are well-suited for our task of fine-scale forecasting. While the end goal is operational heatwave forecasting, we forecast 2-meter air temperature as the primary target because heatwave definitions vary across jurisdictions (for example, consecutive hot days, exceedance of absolute thresholds, or percentile-based criteria). Predicting temperature provides a low-level signal that can be post-processed to align with region- or agency-specific heatwave definitions, without retraining the model.

\begin{figure}[t]
  \centering
  \begin{minipage}[t]{0.32\linewidth}
    \centering
    \includegraphics[width=\linewidth]{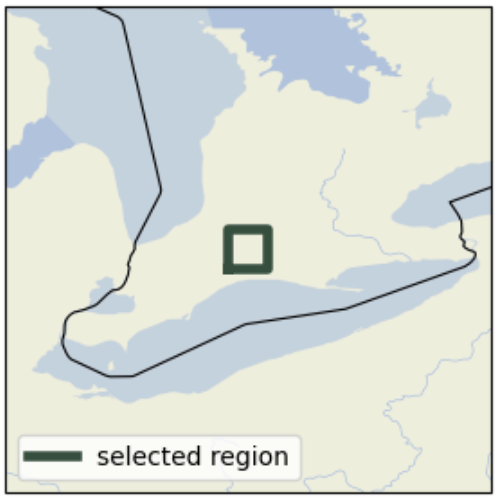}
    \small (a) Region A
  \end{minipage}\hfill
  \begin{minipage}[t]{0.32\linewidth}
    \centering
    \includegraphics[width=\linewidth]{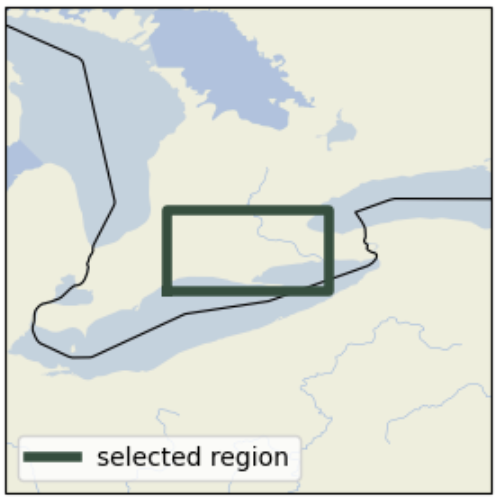}
    \small (b) Region B
  \end{minipage}\hfill
  \begin{minipage}[t]{0.32\linewidth}
    \centering
    \includegraphics[width=\linewidth]{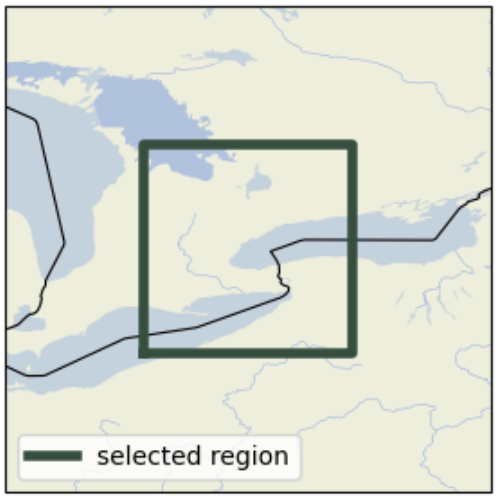}
    \small (c) Region C
  \end{minipage}
  \caption{Bounding boxes around Regions A–C.}
  \label{fig:regions}
\end{figure}

\paragraph{Regions.} Following discussions with stakeholders from climate NGOs in the Global South, we decided to focus on Southwestern Ontario in Canada as our case study region since it incorporates various land types, such as urban, farmland, forest, and water bodies. Specifically, we have 3 bounding boxes in this region, as depicted in Figure~\ref{fig:regions}: \textbf{Region A} is from 42.78\textdegree N, 81.45\textdegree W to 43.18\textdegree N, 81.05\textdegree W ($\approx$ 44~km by 33~km); \textbf{Region B} is from 42.50\textdegree N, 81.50\textdegree W to 43.50\textdegree N, 79.50\textdegree W ($\approx$111~km by 163~km); \textbf{Region C} is from 42.00\textdegree N, 81.00\textdegree W to 45.00\textdegree N, 78.00\textdegree W ($\approx$333~km by 243~km). This setup allows us to run computationally intensive experiments on the smallest window (A), then demonstrate scalability and proof of concept on larger domains with consistent data and preprocessing. All three regions are covered by the URMA analysis grids for the variables of interest.

\paragraph{Preprocessing.} We obtained hourly data for all variables spanning multiple years, with exact ranges specified in the \textit{Appendix}, for each region. Missing or invalid entries were filled using spatial interpolation (i.e., mean of nearby grid points). Each input feature was standardized using the mean and standard deviation of the training set. The target variable (temperature) was also normalized during model training, though final errors are reported in physical units (\textdegree C) for clarity. 

\paragraph{Proposed model.} We developed and trained a hybrid Graph Convolutional Network (GCN) with a Gated Recurrent Unit (GRU) for each region. Each grid point in the region is represented as a graph node with meteorological features (temperature, winds, pressure, etc.), connected by edges to capture spatial interactions. Graph convolution layers model neighborhood effects \cite{kipf2017semi}, while GRUs capture temporal dependencies \cite{cho2014learning}. Train/validation/test splits varied by pipeline and region (specified in the \textit{Appendix}). The objective was to predict temperature 1-48 hours ahead from the current time (at 1, 6, 12, 18, 24, 36, and 48h). The models were optimized using mean squared error (MSE).

\paragraph{Embeddings.} Data from resource-limited regions is often sparse, inconsistent, and difficult to align with information-rich datasets. In a separate setup, to harmonize heterogeneous inputs, we also explore language-model embeddings as an intermediate representation. We convert each Region~A observation (per timestamp and location) into a short paragraph, for example: \textit{\small{``temperature is 291.6 K, dew point is 283.7 K, u~wind component is 4.0 m/s, v~wind component is -2.1 m/s, surface pressure is 99209 Pa, ... , elevation is 172.0 meters.''}} These descriptions are then encoded using a ClimateBERT model \cite{climatebert2022}, yielding a 768-dimensional vector. To control dimensionality and reduce noise, we apply Principal Component Analysis (PCA) \cite{pca}, fitting on train and transforming train, validation, and test. The reduced embeddings are then used as node features within the GCN--GRU forecasting pipeline.

\section{Results}
As summarized in Table~\ref{tab:regions_perf}, the GNNs achieved strong performance across all three regions. Performance improved as the spatial window expanded (Region~C > Region~B > Region~A), which is consistent with larger graphs capturing richer neighborhood interactions and mesoscale context. These results indicate that graph-based models can provide accurate, high-resolution forecasts suitable for downstream early-warning pipelines.

\begin{table}[t]
\centering
\caption{Per-region performance. Mean across horizons; 48h at the farthest forecast horizon. (MAE~=~Mean Absolute Error, RMSE = Root Mean Squared Error)}
\label{tab:regions_perf}
\setlength{\tabcolsep}{5pt}
    \begin{tabular}{lccc}
    \toprule
    Region & Mean MAE (\textdegree C) & MAE@48h (\textdegree C) & RMSE@48h (\textdegree C) \\
    \midrule
    A                    &  2.55 & 3.78  & 4.84 \\
    B                    & 2.48 & 3.73 & 4.84 \\
    C                    & 1.93 & 2.93 & 3.90 \\
    \bottomrule
    \end{tabular}
\end{table}

\begin{table}[t]
\centering
\caption{Region A performance comparisons. Mean across horizons; 48h at the farthest horizon.}
\label{tab:regionA_perf}
\setlength{\tabcolsep}{5pt}
    \begin{tabular}{lcccc}
    \toprule
    Model & Mean MAE (\textdegree C) & MAE@48h (\textdegree C) &  RMSE@48h (\textdegree C)\\
    \midrule
    Baseline (tabular)              & 2.55 & 3.78  & 4.84 \\
    Embeddings (ClimateBERT+PCA)    & 3.34 & 4.34  & 5.54 \\
    Control (random weights)        & 9.11 & 8.89  & 10.49   \\
    \bottomrule
    \end{tabular}
\end{table}

On Region~A, as seen in Table~\ref{tab:regionA_perf}, performance with embeddings shows a modest decrease in mean MAE relative to the tabular baseline, while a control model initialized with random weights performs noticeably worse, indicating that the embedding features carry meaningful signals. This approach provides a standardized input format that can accommodate missing or unstandardized variables, which is valuable when extending to data-limited regions. We expect comparable performance with minimal fine-tuning as additional local data becomes available.

\begin{figure}[t]
  \centering
  \begin{minipage}[t]{0.68\linewidth}
    \centering
    \includegraphics[width=\linewidth]{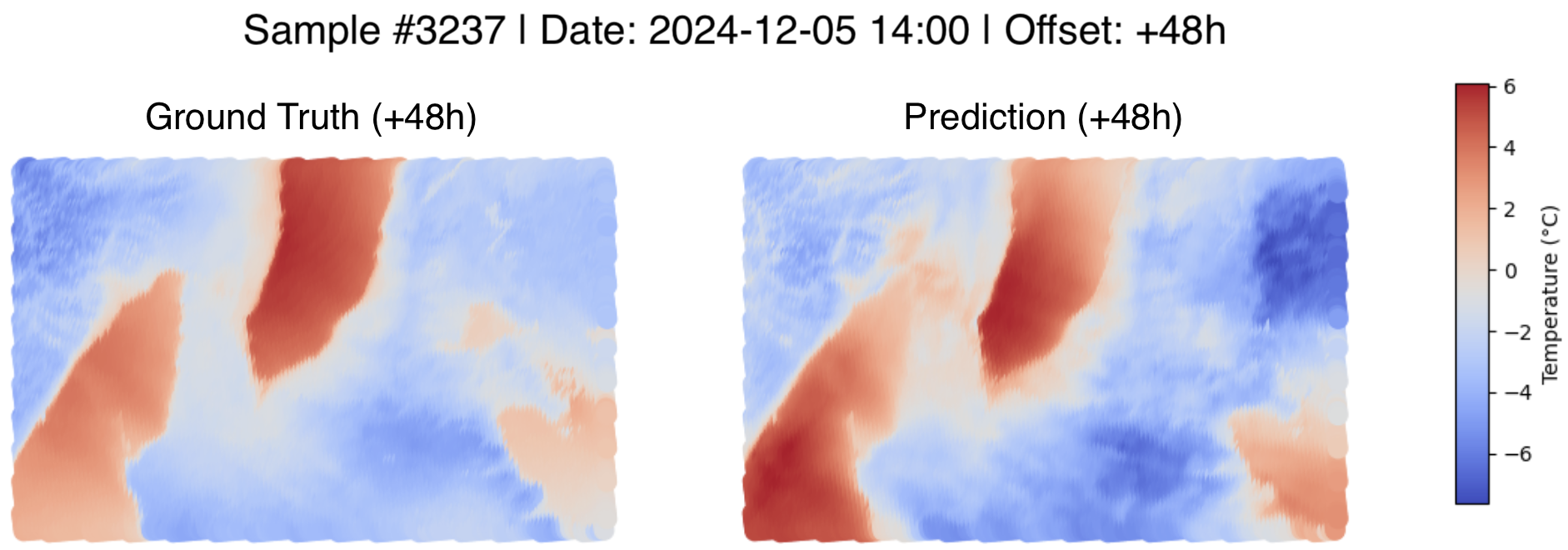}
    \vspace{2pt}\par\small (a) Randomly sampled test timestamp showing ground truth and model predictions 48 hours ahead in Region C.
  \end{minipage}\hfill
    \begin{minipage}[t]{0.28\linewidth}
    \centering
    \includegraphics[width=\linewidth]{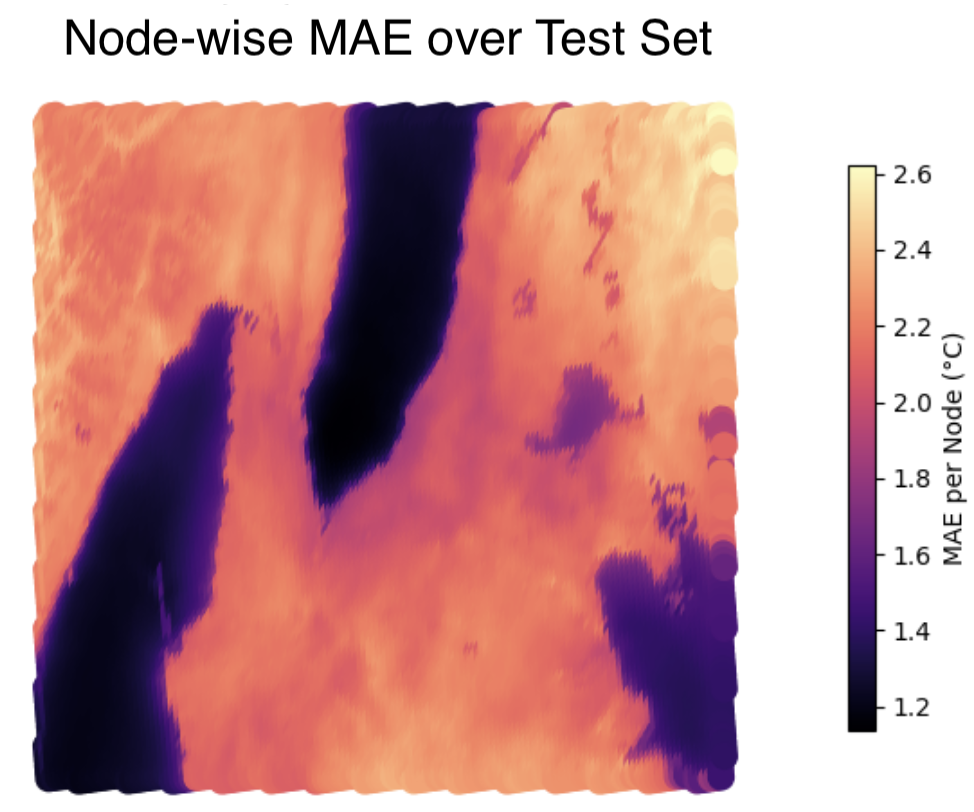}
    \vspace{2pt}\par\small (b) Average node-wise MAE across the test set in Region C.
  \end{minipage}\hfill
 \caption{Example results from Region C.} 
  \label{fig:results}
\end{figure}

\section{Discussion}

Table~\ref{tab:regions_perf} shows a clear trend: MAE and RMSE decrease as region size increases, suggesting that larger graphs capture richer spatial context. Training on Region~C strains memory, so we also sampled every 6\,h. The 6\,h model reached mean MAE~2.39, MAE@48h~3.15, and RMSE@48h~4.16, versus the hourly model’s 1.93, 2.93, and 3.90, respectively---i.e., modest degradations (+0.46, +0.22, +0.26) for substantially lower compute. This indicates that coarser temporal sampling can cut compute demands significantly while retaining most of the forecasting skill. Additionally, the use of embeddings provides a way to standardize non-standard data for modelling.

Although the results are promising, our evaluation is limited to held-out test performance on URMA data. Most benchmarked state-of-the-art systems run at much coarser scales (about 10–50~km, 3–6h), whereas our model operates at 2.5~km and hourly. This resolution mismatch makes fair comparisons difficult and highlights the need for future matched-resolution baselines. Future work should also test transferability to regions with sparser data and different climate regimes.

Localized climate forecasting is not just a technical problem, but an equity issue. Marginalized groups, already at heightened risk due to structural inequalities \cite{smith2022climate}, are further disadvantaged when coarse models fail to detect their specific vulnerabilities \cite{putsoane2024extreme,jessel2019energy}. Our approach offers a transferrable solution: models trained in data-rich regions can be adapted for under-monitored contexts, strengthening forecasting capacity and promoting climate resilience where it is most needed. For Global South governments with limited resources, such forecasts could support early-warning systems that prioritize vulnerable neighborhoods, enabling more efficient allocation of scarce resources (e.g., cooling centers, medical services, energy). By producing forecasts at community scale, GNN-based models can support municipal planning and public health interventions in ways coarse global models cannot.

This work contributes to the dialogue on climate AI \cite{rolnick2022tackling} by showing that small, region-specific GNNs can provide outsized impact compared to large-scale models requiring vast compute. Lightweight, locally deployable systems may offer the most practical path toward equitable climate 
resilience.

\section{Conclusion}
We introduced a GCN--GRU framework for community-scale 2-meter temperature forecasting at 2.5\,km using NOAA URMA across three regions in Southwestern Ontario. The models achieved strong performance, with MAE and RMSE improving as spatial context increased. A lighter 6-hour sampling preserved most of the hourly model’s skill, and a language-model embedding pathway provided a practical route to standardize inputs while maintaining useful performance.

Although the long-term goal is heatwave early warning, forecasting temperature offers a flexible, low-level signal that can be post-processed to match specific heatwave definitions. Future directions include expanding predicted targets beyond temperature to include humidity, wind, and related variables, enabling a composite index and more reliable event detection. To address limitations such as resolution mismatch with coarser baselines and evaluation only on URMA data, we plan matched-resolution baselines for fairer comparisons and broader geographic coverage. Beyond this, integrating with operational dashboards will help the system support timely and equitable early warnings at low computational and environmental cost. The same framework could also extend to other extremes, such as wildfires, floods, or droughts.

\bibliographystyle{unsrt} % from docs to make the number citations in order
\bibliography{tacking_climate_workshop}

\pagebreak
\appendix
\section*{Appendix}

\begin{table}[h!]
\centering
\caption{Description of selected NOAA URMA variables.}
\label{tab:urma_vars}
    \begin{tabular}{ll}
    \toprule
    \textbf{Abbreviation} & \textbf{Definition} \\
    \midrule
    t2m   & 2-meter air temperature (K) \\
    d2m   & 2-meter dewpoint temperature (K) \\
    u10   & 10-meter u-component of wind (m/s) \\
    v10   & 10-meter v-component of wind (m/s) \\
    sp    & Surface pressure (Pa) \\
    orog  & Orography (m) \\
    \bottomrule
    \end{tabular}
\end{table}

\begin{table}[h!]
\centering
\caption{Baseline experimental setup: data frequency and splits.}
\label{tab:baseline_splits}
\setlength{\tabcolsep}{5pt}
\begin{tabular}{lccccccc}
\toprule
Region & Frequency & Start Date & End Date & Split & Train;Val;Test Periods/Timestamps \\
\midrule
A   & 1hr & 01/2022 & 06/2025 & Manual & 2022 ; 2023 ; 2024-end \\
B   & 1hr & 01/2022 & 06/2025 & Manual & 2022 ; 2023 ; 2024-end  \\
C   & 1hr & 01/2022 & 12/2024 & Ratio & 18,396 ; 3,942 ; 3,943 \\
C (6hr)  & 6hr & 06/2017 & 06/2025 & Ratio & 8,170 ; 1,751 ; 1,751 \\
\bottomrule
\end{tabular}
\par\addvspace{5pt} 
    \noindent{\footnotesize
      \textbf{Additional information:} Features included: t2m, d2m, u10, v10, sp, orog.
      Splits were chosen based on the best-performing models identified via grid search.  
      For regions with data extending to June 2025 (A and B), we included these months in the test set rather than shifting timestamp boundaries, to avoid bias from introducing additional seasonal data.  
      Ratio-based splits followed a 70\%/15\%/15\% scheme.  
      For Region C, due to resource limitations, we excluded the additional six months available in Regions A and B that were extra for testing.  
    }
\end{table}

\begin{table}[h!]
\centering
\caption{Baseline experimental setup: hyperparameters and forecast horizons.}
\label{tab:baseline_hyperparams}
\setlength{\tabcolsep}{5pt}
\begin{tabular}{lccc}
\toprule
Region & Hyperparameters & Forecast Horizons (hrs) \\
\midrule
A   & LR=0.0001, Win=48, BS=16, HD=32, Dist=8 & 1, 6, 12, 18, 24, 36, 48 \\
B & LR=0.001, Win=24, BS=16, HD=32, Dist=8 & 1, 6, 12, 18, 24, 36, 48 \\
C & LR=0.001, Win=24, BS=16, HD=32, Dist=4 & 1, 6, 12, 18, 24, 36, 48 \\
C (6hr)& LR=0.001, Win=24, BS=16, HD=32, Dist=4 & 6, 12, 18, 24, 36, 48 \\
\bottomrule
\end{tabular}
\par\addvspace{5pt} 
    \noindent{\footnotesize
      \textbf{Abbreviations:} LR = learning rate; Win = input window (hrs), BS = batch size; HD = hidden dimensions; Dist~= distance (threshold for graph connectivity (km)). Final hyperparameters were selected based on the best-performing models identified via grid search.  
    }
\end{table}

\begin{table}[h!]
\centering
\caption{Training setup: GPU hardware, number of GPUs, and runtime per region.}
\label{tab:training_setup}
\setlength{\tabcolsep}{5pt}
\begin{tabular}{lccc}
\toprule
Region & GPU Model & \#GPUs & Runtime \\
\midrule
A   & NVIDIA L40S (48GB) & 2 & 35 min \\
B   & NVIDIA L40S (48GB) & 2 & 42 min \\
C   & NVIDIA H100 SXM (80GB) & 2 & 5h 40 min \\
C (6hr) & NVIDIA L40S (48GB) & 2 & 5h 45 min \\
\bottomrule
\end{tabular}
\par\addvspace{5pt}
    \noindent{\footnotesize
      \textbf{Cluster specs:} L40S nodes - Dell 750xa, 2$\times$Intel Xeon Gold 6338, 512 GB RAM, 4$\times$NVIDIA L40S 48GB.  
      H100 nodes - Dell XE9680, 2$\times$Intel Xeon Gold 6442Y, 2048 GB RAM, 8$\times$NVIDIA H100 SXM 80GB.  
    }
\end{table}

\end{document}